\documentclass[10pt,twocolumn,letterpaper]{article}

\usepackage{wacv}
\usepackage{times}
\usepackage{epsfig}
\usepackage{graphicx}
\usepackage{amsmath}
\usepackage{amssymb}

\usepackage{float}

\wacvfinalcopy

\ifwacvfinal\fi
\begin{document}

\title{Generalised Zero-Shot Learning with a Classifier Ensemble over Multi-Modal Embedding Spaces}

\author{Rafael Felix$^{1,2}$, Ben Harwood$^{1,3}$, Michele Sasdelli$^{1,2}$ and Gustavo Carneiro$^{1,2}$ \\
Australian Centre for Robotic Vision$^{1}$\\
The University of Adelaide$^{2}$\\
Monash University$^{3}$\\
{\tt\small \{rafael.felixalves,michele.sasdelli,gustavo.carneiro\}@adelaide.edu.au, ben.harwood@monash.edu}
}

\maketitle
\ifwacvfinal\thispagestyle{empty}\fi

\begin{abstract}
Generalised zero-shot learning (GZSL) methods aim to classify previously seen and unseen visual classes by leveraging the semantic information of those classes. In the context of GZSL, semantic information is non-visual data such as a text description of both seen and unseen classes. Previous GZSL methods have utilised transformations between visual and semantic embedding spaces, as well as the learning of joint spaces that include both visual and semantic information. In either case, classification is then performed on a single learned space. We argue that each embedding space contains complementary information for the GZSL problem. By using just a visual, semantic or joint space some of this information will invariably be lost. In this paper, we demonstrate the advantages of our new GZSL method that combines classification of visual, semantic and joint spaces. Most importantly, this ensembling allows for more information from the source domains to be seen during classification. An additional contribution of our work is the application of a calibration procedure for each classifier in the ensemble. This calibration mitigates the problem of model selection when combining the classifiers. Lastly, our proposed method achieves state-of-the-art results on the CUB, AWA1 and AWA2 benchmark data sets and provides competitive performance on the SUN data set.

\end{abstract}

\section{Introduction}\label{sec:introduction}

Visual classification is the process of assigning class labels to images or image regions in a way that accommodates invariances such as pose deformation, changes in lighting and differences in the appearance of individual samples. Classification systems that operate over a broad range of classes will typically learn visual features from a labelled training set and then detect those features in previously unseen test sets. These features should be invariant to intra-class variations, while also representing discriminating inter-class variations. Generalised zero-shot learning (GZSL) is a much harder formulation of this classification task~\cite{xian2017zero}. In GZSL, the test set is augmented with samples from unseen classes that were not present in the training set. As such, any class variations being modelled by the learned features should be as general as possible. To assist with this generalisation, GZSL methods currently make use of an auxiliary training set that contains semantic information about all classes (i.e. text descriptions for the seen classes in the training/test sets and the unseen classes in the test set). 

Traditional GZSL methods aim to build a function that transforms samples from the visual to the semantic space so that the classification of seen and unseen classes are performed exclusively in the semantic space~\cite{xian2017zero}.  More recent approaches, such as the adaptive confidence smoothing (COSMO)~\cite{atzmon2019adaptive}, classify on a single visual feature space. Here, a classifier for seen classes is trained on the training images, while a classifier for unseen classes is trained with synthetic visual features that are generated from the semantic data set. Similarly, cycle-WGAN~\cite{felix2018multi} trains a generative adversarial network (GAN) that generates synthetic visual features from the semantic data set. These new features are then used to augment the original training data set and train a GZSL classifier for the seen and unseen classes. In contrast, CADA-VAE~\cite{schonfeld2019cada} encodes the semantic data directly into the same embedding space as the visual data to create a joint semantic/visual space. This latent encoding is then refined by learning to decode the latent embedding vectors to both of the input domains. The joint space is then used for the classification of all classes. Ultimately, each of these methods bottlenecks the information presented to the classifier within a single learned embedding space. This is potentially at odds with a key challenge of GZSL, in that it is not known ahead of time exactly what training data will be useful for classifying the unseen classes.

\begin{figure*}[ht!]
    \centering
    \includegraphics[width=13cm]{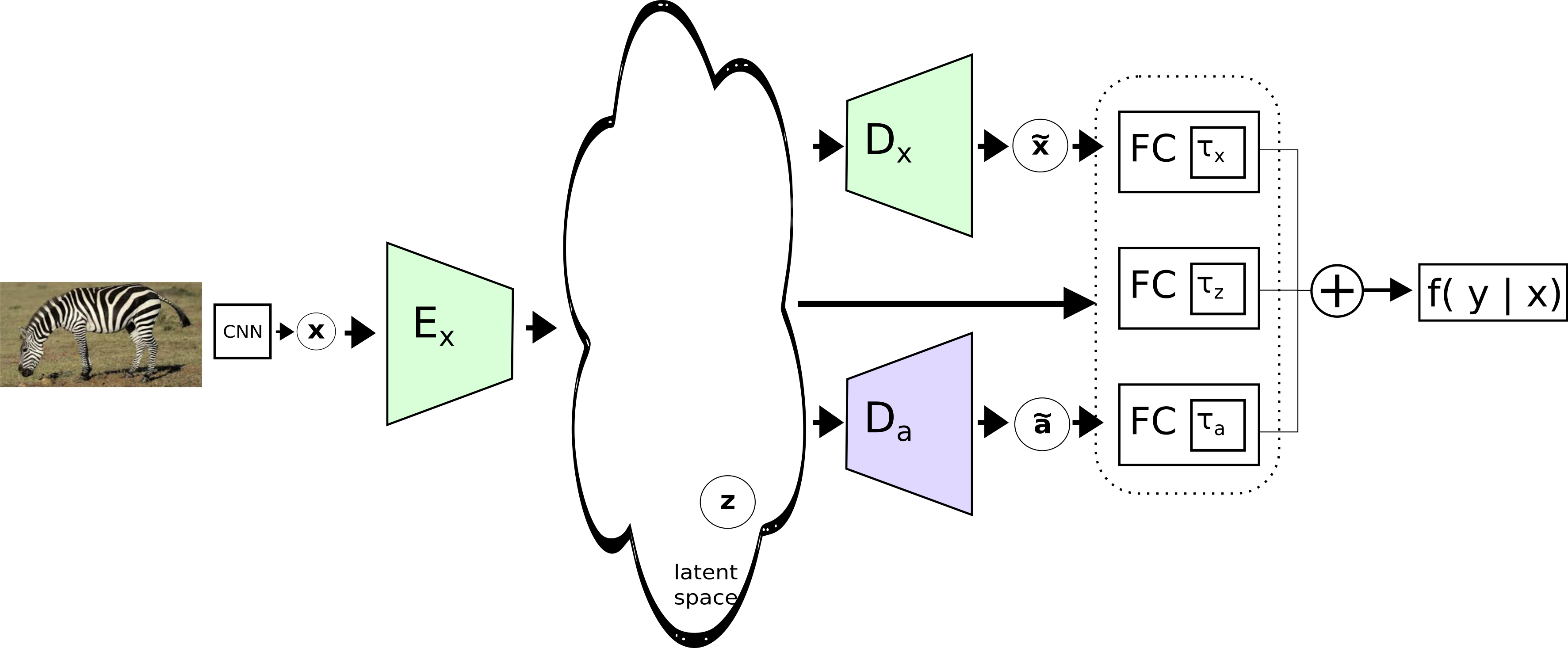}
    \caption{Depiction of our ensemble of GZSL classifiers. In testing time, a image is presented to our pipeline. Then, the visual representation is transformed to the shared visual/semantic latent space, represented by $\textbf{z}$. This latent vector is then transformed to the visual ($\tilde{\textbf{x}}$) and semantic ($\tilde{\textbf{a}}$) spaces. These multi-modal embedding vectors, ($\tilde{z}$, $\tilde{\textbf{x}}$, and $\tilde{\textbf{a}}$), are used to produce the classification of a test image, coming from seen or unseen classes.}
    \label{fig:motivation}
\end{figure*}

In this paper, we demonstrate the benefits of a new GZSL method, consisting of an ensemble of classifiers that achieves better performance on the GZSL problem. As with some existing GZSL methods~\cite{schonfeld2019cada}, we learn a single embedding space that is shaped by both the visual and semantic training data. However, our method also leverages the additional information present in the learned mappings back to each of the source domains. As depicted in Fig.~\ref{fig:motivation}, we achieve this by classifying with an ensemble of our learned joint embedding space, a reconstructed visual space and a reconstructed semantic space. Then for each of our three embedding spaces, we apply a neural network temperature scaling calibration technique. Although calibration of neural networks has been commonly adopted in supervised learning methods~\cite{guo2017calibration}, there are currently no GZSL methods that have adopted this approach. Incidentally, we demonstrate that this calibration also mitigates the problem of GZSL model selection when combining the classification from each embedding space.

The main body of this paper is divided into two main components. In Sections~\ref{sec:literature} and~\ref{sec:method} we contextualise the GZSL problem and detail the key components of our proposed approach. Then in Sections~\ref{sec:experiments} and~\ref{sec:discussions} we provide specific implementation details, results and discussions of experiments run with our GZSL method. In particular, this includes an ablation study that highlights the structural differences between each of the three feature spaces considered by our classifier. We also show that our ensemble consistently improves upon the harmonic-mean and unseen top-1 accuracy of baseline CADA-VAE results across various data sets. With these improvements, our approach sets new state-of-the-art classification accuracy for three of the four data sets, commonly used in GZSL problems.

\section{Literature Review}\label{sec:literature}

In this section, we review the recent literature that is relevant to our work.  We first introduce zero-shot learning, then describe generalised zero-shot learning, providing details on data augmentation for GZSL, and another section on ensemble methods for GZSL classification. 

\subsection{Zero-Shot Learning}

Zero-shot learning (ZSL) is defined as a classification problem, where the set of (seen) visual classes used for training does not overlap with the set of (unseen) visual classes used for testing~\cite{lampert2014attribute,xian2017zero}.  The main solution explored by ZSL methods is based on the use of an auxiliary semantic space, where each (seen and unseen) visual class has a particular semantic representation.  With the learning of a transformation function that projects from visual to semantic representations, it is then possible to transform samples from unseen visual classes to the semantic space, where the assumption is that this projection should lie close to its correct semantic representation.  This set-up limits the applicability of ZSL methods~\cite{xian2017feature,felix2018multi} because it is important that the method should be able to recognise not only the unseen but also the seen visual classes during testing~\cite{chao2016empirical,changpinyo2016synthesized}.  Although limited, ZSL methods can be seen as an expert model for the unseen visual classes~\cite{atzmon2019adaptive}.

\subsection{Generalized Zero-Shot Learning}
\label{sec:GZSL_litreview}

GZSL extends the ZSL framework with the recognition of the seen and unseen visual classes during testing.  Such an extension is troublesome because of the bias of the learned classifier toward the seen classes.  That is, a new test sample from a seen or unseen class is more likely to be classified as one of the seen classes than one of the unseen classes. This issue is one of the main motivations behind the development of several GZSL approaches~\cite{xian2017feature}.  For instance, anomaly detection can be used to identify a visual sample from an unseen class~\cite{socher2013zero} that can then be handled differently from a seen class sample.  Similarly, an auxiliary seen versus unseen domain classifier can be used to reduce the bias toward seen classes~\cite{chao2016empirical}.  

The most successful GZSL approaches are based on methods that generate synthetic visual samples for the unseen classes, given their semantic representation~\cite{felix2018multi,schonfeld2019cada,xian2017feature,verma2018generalized}.  These artificial unseen class samples, together with the real seen visual samples, are used to train a visual classifier of seen and unseen classes.  The generative model explored by these methods is the Generative Adversarial Networks (GAN)~\cite{felix2018multi,xian2017feature} and Variational Autoencoders (VAE)~\cite{schonfeld2019cada,verma2018generalized}.  The approaches above do not have a testing stage that can handle the multi-modal (i.e., visual and semantic) nature of the GZSL classification -- in fact, during the testing stage, these approaches only deal with samples either in the visual space or in a joint visual and semantic space.  We, on the other hand, target the use of all spaces available during the testing stage: the reconstructed visual, reconstructed semantic, and joint visual and semantic spaces. 

The approach developed by Atzmon and Chechik~\cite{atzmon2019adaptive} also targets the bias issue toward seen classes.  Their solution involves a classifier that combines the result of a ZSL classifier (for the unseen classes) and a seen class classifier, where such combination is achieved with a (seen/unseen) gating network. Even though such an approach achieves outstanding results, it can be criticised for not exploring more effectively the multi-modality nature of the problem and for relying on models that are sub-optimally trained with subsets of the data set.  Our approach aims to optimally combine multi-modal classifiers in a solution that effectively mitigates the issue of training a seen/unseen domain classifier.

\subsection{Ensemble of Calibrated Classifiers}

Ensemble of classifiers is a widely used technique for boosting classification performance in several problems~\cite{dietterich2000ensemble}.  In ZSL and GZSL, the ensemble of classifiers has been applied to combine semantic classifiers~\cite{8100119}, to learn transformations between the visual and semantic spaces~\cite{zhang2018triple} and to combine the seen and unseen classifications~\cite{atzmon2019adaptive,chao2016empirical}.  Even though the ensemble of classifiers is a powerful method, the actual combination generally involves hyper-parameters to weight the contribution of each classifier, which requires non-trivial training procedures that tend to rely on model selection based on cross-validation.  In addition, none of the GZSL methods explicitly explore a multi-modal ensemble of classifiers.

In our paper, we take a different path to ensemble multi-modal classifiers that automatically handles seen versus unseen domain classification.  More specifically, we rely on an end-to-end training of multi-modal visual, semantic, and joint visual semantic classifiers that are trivially combined by averaging their classification results.  We show that such trivial ensemble is possible because we calibrate the classification result~\cite{guo2017calibration} for each modality. Classification calibration consists of a post-processing method that ensures that the classification probability correlates well with the classifier output -- this post-processing learns a single parameter per classifier and does not change the classification accuracy.

\section{Method}\label{sec:method}
In this section, we first introduce the GZSL problem and a data augmentation framework for GZSL. Then, we discuss the proposed ensemble of models. Finally, we introduce the calibration of GZSL classifiers.

\subsection{Generalised Zero-Shot Learning}\label{sec:problem_formulation}

GZSL methods rely on two data modalities. The data set for the visual modality is represented as $\mathcal{D} = \{(\textbf{x},  y)_i\}_{i=1}^{N}$, where $\textbf{x} \in \mathcal{X} \subseteq \mathbb{R}^X$ denotes the visual representation, and $y \in \mathcal{Y} = \{ 1,..., C \}$ denotes the visual class. The visual representation consists of visual features extracted by pre-trained deep neural networks, such as ResNet~\cite{he2016resnet}, and VGG~\cite{simonyan2014very}. In GZSL problems, $\mathcal{D}$ is split into two disjoint domains: the seen domain $\mathcal{Y}^S = \{1, ... , |S|\}$, and the unseen domain $\mathcal{Y}^U = \{(|S|+1), ... , (|S|+|U|)\}$, where $\mathcal{Y} = \mathcal{Y}^S \cup \mathcal{Y}^U$, and $\mathcal{Y}^S \cap \mathcal{Y}^U = \emptyset$. 
Samples from $\mathcal{Y}^S$ can be accessed during training time, but samples from the unseen domain $\mathcal{Y}^{U}$ are only available during test time. Therefore the main challenge in GZSL consists of classifying samples that are drawn from $\mathcal{Y}$, independently if they come from the seen or unseen domain~\cite{xian2017zero}.
The data set for the semantic modality is defined as $\mathcal{R} = \{ (\mathbf{a},y)_j \}_{j \in \mathcal{Y}}$, where each $\mathbf{a} \in \mathcal{A} \subseteq \mathbb R^A$ is associated to a visual class from $\mathcal{Y}$. The semantic representation consists of a semantic information (e.g., textual description, or a set of attributes) available for the visual classes. This information can be projected into a embedding space by feature representation methods (e.g., set of continuous features such as \textit{word2vec}~\cite{xian2017zero}, or \textit{BoW}). In this problem, we define that the semantic data set has only one representation per class.

GZSL has a particular set up for the training and testing stages.  The data set $\mathcal{D}$ is divided into two subsets: $\mathcal{D}^{tr}$ for training, and $\mathcal{D}^{ts}$ for testing. The training set contains visual samples drawn from the seen classes $\mathcal{Y}^S$ and the testing set contains samples from both the seen and unseen domains. The semantic data set, $\mathcal{R}$, is available during training and testing.

\subsection{Data Augmentation Framework}
\label{sec:data_augmentation}

\begin{figure}[h!]
    \centering
    \includegraphics[width=3.25in,height=2in]{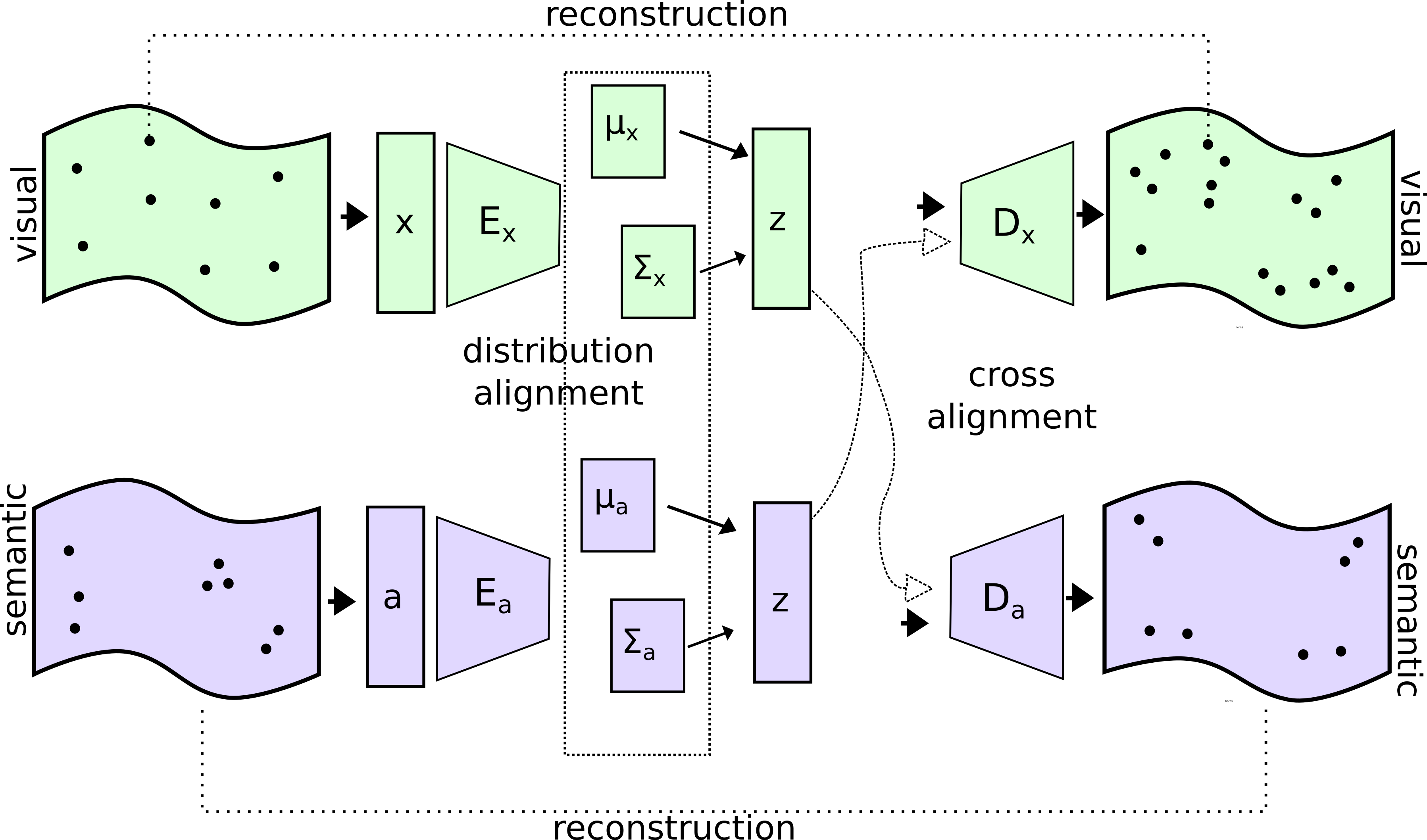}
    \caption{Depiction of CADA-VAE~\cite{schonfeld2019cada}. The encoder networks for the visual and semantic representation transform samples from each modality into a joint visual/semantic latent space via VAEs optimised with cross-modality alignment and distribution alignment, as explained in~\eqref{eq:cada_da}.}
    \label{fig:cada_vae}
\end{figure}

There have been many GZSL methods that relies on the generation of artificial visual samples, given their semantic representation~\cite{felix2018multi,schonfeld2019cada,xian2017feature,verma2018generalized}, as described in Sec.~\ref{sec:GZSL_litreview} -- these methods are referred to as data augmentation framework approaches.  In this paper, we extend the model CADA-VAE~\cite{schonfeld2019cada} given its state-of-the-art GZSL classification results. CADA-VAE~\cite{schonfeld2019cada} uses a variational auto-encoder~\cite{kingma2013auto} that consists of a visual encoder $E_{x}: \mathcal{X} \rightarrow \mathcal{Z}$ (where $\mathcal{Z} \in \mathbb R^Z$ denotes the latent joint visual/semantic space), a semantic encoder $E_{a}: \mathcal{A} \rightarrow \mathcal{Z}$, a visual decoder $D_{x}: \mathcal{Z} \rightarrow \mathcal{X}$ and a semantic decoder  $D_{a}: \mathcal{Z} \rightarrow \mathcal{A}$.  
The first stage of CADA-VAE training estimates the latent space with the following optimisation loss (Fig.~\ref{fig:cada_vae}):
\begin{equation}
\begin{split}
    \mathcal{L} = \mathcal{L}_{VAE} + \gamma_{CM}~\mathcal{L}_{CM} + \gamma_{DA}~\mathcal{L}_{DA},
\end{split}
\label{eq:cada_vae}
\end{equation}
where the first term represents the VAE loss~\cite{schonfeld2019cada}, the second denotes the cross-modality alignment loss that calculates the reconstruction error between the visual and semantic modalities, defined by
\begin{equation}
\begin{split}
    \mathcal{L}_{CM} =  \sum_i \| \mathbf{x}_i - D_{x}(E_{a}(\mathbf{a}_i))  \| + \| \mathbf{a}_i - D_{a}(E_{x}(\mathbf{x}_i))  \|.
\end{split}
\label{eq:cada_ca}
\end{equation}
The last term in~\eqref{eq:cada_vae}, $\mathcal{L}_{DA}$, consists of the distribution-alignment loss of samples belonging to the same class, defined by
\begin{equation}
\begin{split}
    \mathcal{L}_{DA} = \sum_i \mid\mid \mu_{(\mathbf{x}_i)} - \mu_{(\mathbf{a}_i)}\mid\mid_2^2 
                + \mid\mid \Sigma_{(\mathbf{x}_i)}^{\frac{1}{2}} - \Sigma_{(\mathbf{a}_i)}^{\frac{1}{2}}\mid\mid_{F}^2,
\end{split}
\label{eq:cada_da}
\end{equation}
where $\mu_{(\mathbf{x}_i)} \in \mathcal{Z}$ and $\Sigma_{(\mathbf{x}_i)} \in \mathcal{Z} \times \mathcal{Z}$ are the mean vector and co-variance matrix of the latent samples from a particular class produced by the encoder $E_{x}(.)$ \big(similarly for $\mu_{(\mathbf{a}_i)}$ and $\Sigma_{(\mathbf{a}_i)}$ for $E_{a}(.)$\big), and $F$ represents the Frobenius norm.  This loss assumes a uni-modal Gaussian distribution of the latent vectors of a particular class, and approximates the distributions produced by the visual and semantic classes (Fig.~\ref{fig:cada_vae}).

This first stage of training will build two encoders and two decoders, where it is then possible to encode latent samples in the joint visual/semantic space for both the seen and unseen classes.  Then the second training stage of CADA-VAE consists of using these samples to build a classifier for the seen and unseen classes, which is denoted by $p(y | \mathbf{z})$, where $\mathbf{z} \in \mathcal{Z}$ can come from $E_{x}(.)$ or $E_{a}(.)$.  

\subsection{Ensemble of Multi-modal Classifiers}
\label{sec:ensemble_MM_classifiers}

The main contribution of this paper is the combination of the three modalities for GZSL classification, as follows:
\begin{equation}
\begin{split}
p(y | \mathbf{x}) = & p(y | E_{x}(\mathbf{x})) + \lambda_x p(y | D_{x}(E_{x}(\mathbf{x})) + \\
& \lambda_a p(y | D_{a}(E_{x}(\mathbf{x})).
\end{split}
\label{eq:proposed_multimodal_classifier}
\end{equation}
The training for the multi-modal classifier in~\eqref{eq:proposed_multimodal_classifier} is based on optimising the negative log-likelihood for each of of the classifiers. One important issue is the estimation of the weights $\lambda_x$ and $\lambda_a$ that usually involves a time-consuming model selection process based on cross validation.  We avoid such process by simply calibrating each of the multi-modal classifiers with temperature scaling~\cite{guo2017calibration}, which, for the case of softmax classifier, is defined by:
\begin{equation}
    \label{eq:softmax_temperature}
    p(  y \mid \mathbf{z}) = \frac{{\rm e}^{(\pi(y | \mathbf{z})/\tau)}}
    {\sum_{c=1}^{C} {\rm e}^{(\pi(y = c|\mathbf{z})/ \tau)}},
\end{equation}
where $\mathbf{z} = E_{x}(\mathbf{x})$, and $\pi(y_i|\mathbf{z})$ represents the output logits for $p(  y \mid \mathbf{z})$ (and similarly for $p(y | D_{x}(E_{x}(\mathbf{x}))$ and $p(y | D_{a}(E_{x}(\mathbf{x}))$ in~\eqref{eq:proposed_multimodal_classifier}). In traditional supervised learning, the temperature scaling factor $\tau$ is assumed to be equal to one. However, recent research shows that this parameter can be used for calibrating the classification confidence~\cite{guo2017calibration}.
After calibrating each classifier, the ensemble consists of an averaging of the three classification results from~\eqref{eq:proposed_multimodal_classifier}.
The testing procedure for the classifier in~\eqref{eq:proposed_multimodal_classifier} involves taking a testing visual sample $\mathbf{x}$ and classify it from the average of the three multi-modal classifiers from~\eqref{eq:proposed_multimodal_classifier}.

\section{Experiments}
\label{sec:experiments}

In this section, we introduce the experimental setup to demonstrate the performance of the proposed method. First, we present the benchmark data sets, then we describe the evaluation criteria for our experimental setup.  We have a section to show the results of our method compared with the current state-of-the-art. Finally, we provide ablation studies to explore the functionality of our method.

\begin{table*}[ht!]
\centering
\label{table:dataset-stats}
\caption{The benchmarks for GZSL: CUB\cite{welinder2010caltech}, SUN \cite{xiao2010sun}, AWA1\cite{xian2017zero}, and AWA2~\cite{xian2017zero}. Column (1) shows the number of seen classes, denoted by $|\mathcal{Y}^S|$, split into the number of training and validation classes (train+val), (2) presents the number of unseen classes $| \mathcal{Y}^U |$, (3) displays the number of samples available for training $|\mathcal{D}^{Tr}|$ and (4) shows number of testing samples that belong to the unseen classes $|\mathcal{D}_U^{Te}|$ and number of testing samples that belong to the seen classes $|\mathcal{D}_S^{Te}|$ from~\cite{felix2018multi,xian2017feature}}
\begin{tabular}{|l|c|c|c|c|}
\hline
\textbf{Name} & $|\mathcal{Y}^S|$ (train+val) & $|\mathcal{Y}^U|$ & $|\mathcal{D}^{Tr}|$ & $|\mathcal{D}^{Te}_U|+|\mathcal{D}^{Te}_S|$
\\
\hline
\hline
CUB     & 150 (100+50) & 50 & 7057 & 1764+2967 \\
SUN    & 745 (580+65) & 72 & 14340 & 2580+1440\\
AWA1    & 40 (27+13) & 10 & 19832 & 4958+5685\\
AWA2 & 40 (27+13) & 10 & 23527 & 5882+7913\\
\hline
\end{tabular}
\end{table*}

\subsection{Data Sets}

We evaluate our method on four publicly available\footnote{Datasets from https://cvml.ist.ac.at/AwA2/.} benchmark GZSL data sets: SUN~\cite{xian2017zero}; AWA1~\cite{lampert2009lerning,xian2017zero}, AWA2~\cite{lampert2009lerning,xian2017zero} CUB-200-2011~\cite{welinder2010caltech}. Recent research argues that GZSL approaches that use pre-trained models must take into consideration the overlap between unseen classes and the ImageNet classes~\cite{xian2017zero}. Therefore, we use the GZSL experimental setup described by Xian et al.~\cite{xian2017zero}, which prevents that the GZSL unseen classes overlap with the ImageNet classes. Moreover, we evaluate our model in terms of fine and coarse-grained data sets. Firstly, we describe the Birds-CUB-200-2011~\cite{welinder2010caltech} as a fine-grained data set. The samples in a fine-grained data set are visually similar to each other, and the semantic representation is generally discriminative with specific details. On the other hand, the data sets SUN, AWA1 and AWA2 are coarse-grained data sets. In particular, the data set SUN represents a challenging problem to GZSL due to the number of classes, and the diversity of these classes~\cite{xian2017zero}. Table~\ref{table:dataset-stats} contains some basic information about the data sets in terms of the number of seen and unseen classes and the number of training and testing images.

\subsection{Feature Representation}

The visual representation for all the benchmark data sets is extracted from the activation of the 2048-dimensional top pooling layer of ResNet-101~\cite{he2016resnet}. The semantic representation of CUB-200-2011~\cite{xian2017zero} consists of the 1024-dimensional vector produced by CNN-RNN~\cite{reed2016learning}. These semantic samples represent a written description of each image using 10 sentences per image. To define a unique semantic sample per-class, we average the semantic samples of all images belonging to each class~\cite{xian2017zero}. For AWA1, AWA2 and SUN we used the proposed semantic features proposed by ~\cite{xian2017zero}. In particular, we use the 102, 85 and 85-dimensional feature space for the data sets SUN~\cite{xian2017zero}, AWA1~\cite{xian2017zero}, and AWA2~\cite{xian2017zero}. 

\subsection{Evaluation Protocol}
\label{sec:evaluation_protocol}

We evaluate our proposed model with the protocol defined by Xian et. al ~\cite{xian2017zero}. Recent research shows that this evaluation protocol has been widely used for GZSL evaluation ~\cite{xian2017zero,xian2017feature,felix2018multi}. The protocol defined by Xian et. al.~\cite{xian2017zero} relies on three measures: top-1 accuracy for the seen samples, top-1 accuracy for the unseen samples, and the harmonic mean. The top-1 accuracy is computed by the average per-class, then we calculate the overall mean. We calculate the mean-class accuracy for each domain separately, i.e., the seen ($\mathcal{Y}^S$) and the unseen ($\mathcal{Y}^U$) classes. The harmonic mean (H-mean) is a measure that combines the accuracy for the seen and unseen domain~\cite{xian2017zero}.We also present experiments using the area under the seen and unseen curve (AUSUC)~\cite{chao2016empirical}. The AUSUC consists of an experiment that relies a variable hyper-parameter that provides a convex combination of the classification between the seen and the unseen classes~\cite{chao2016empirical}.  The AUSUC is a similar evaluation metric to harmonic mean, which aims to measure the trade-off between seen and unseen domains.

\subsection{Implementation Details}

\begin{table*}[h!]
\centering
\caption{Ablation studies of our GZSL approach, following the same structure displayed in Table~\ref{table:gzsl_results}. We report the results for each of the embedding spaces used for classification, the simple average combination without classification calibration (denoted as $\tau = 1$), and the proposed temperature calibrated ensemble method MCADA-VAE.}
\label{table:gzsl_ablation}
\centering

\resizebox{\textwidth}{!}{
\begin{tabular}{|l|lll|lll|lll|lll|}
\hline
& & \textbf{CUB} & 
& & \textbf{SUN} &
& & \textbf{AWA1} &
& & \textbf{AWA2} &
\\
\textbf{Classifier}  
&  $\mathcal{Y}^S$ & $\mathcal{Y}^U$ & $H$
&  $\mathcal{Y}^S$ & $\mathcal{Y}^U$ & $H$
&  $\mathcal{Y}^S$ & $\mathcal{Y}^U$ & $H$
&  $\mathcal{Y}^S$ & $\mathcal{Y}^U$ & $H$ 
\\ \hline \hline

{$\tilde{\textbf{x}}$-CADA-VAE} \textbf{*}
& $65.0$ & $28.0$ & $39.1$
& $28.9$ & $48.7$ & $36.3$
& $76.5$ & $44.1$ & $56.0$
& $81.4$& $43.8$ & $57.0$ \\

{$\tilde{\textbf{a}}$-CADA-VAE} \textbf{*}
& $61.5$ & $25.0$ & $35.6$
& $24.7$ & $36.7$ & $29.5$
& $77.0$ & $42.1$ & $54.4$
& $81.9$& $47.9$ & $60.4$ \\

{$\textbf{z}$-CADA-VAE} \textbf{*} &
$57.2$ & $48.4$ & $52.4$  & 
$36.8$ & $45.1$ & $40.6$  & 
$76.6$ & $55.0$ & $64.1$  & 
$75.3$ & $55.5$ & $63.9$ \\

{MCADA-VAE ~($\tau=1$)}
& $66.7$ & ${30.1}$ & ${41.5}$
& $32.8$ & $\textbf{49.2}$ & $39.3$
& $80.0$ & ${51.3}$ & ${62.5}$
& $84.4$ & ${52.0}$ & ${64.4}$\\

{MCADA-VAE} (ours)
& $55.2$ & $\textbf{52.7}$ & $\textbf{54.0}$
& $35.6$ & $47.4$ & $\textbf{40.7}$
& $75.2$ & $\textbf{57.3}$ & $\textbf{65.0}$
& $73.2$ & $\textbf{58.5}$ & $\textbf{65.0}$
\\
\hline

\end{tabular}
}
\end{table*}

\begin{table*}[ht!]
\centering
\caption{GZSL results using per-class average top-1 accuracy on the test sets of unseen classes $\mathcal{Y}^U$, seen classes $\mathcal{Y}^S$, and H-mean result $H$; -- all results shown in percentage. The results from previously proposed methods in the field were extracted from~\cite{xian2017zero}. The highlighted values represent the best ones in each column (we do do not show the best results for $\mathcal{Y}^S$ because it essentially displays the bias toward seen classes, explained in Sec.~\ref{sec:GZSL_litreview}). The notation $\textbf{*}$ represents the results that we reproduced, and results represented with $-$ were not available in the literature, or hyper-parameters were not given.}
\label{table:gzsl_results}
\centering

\resizebox{\textwidth}{!}{
\begin{tabular}{|l|lll|lll|lll|lll|}
\hline
& & \textbf{CUB} & 
& & \textbf{SUN} &
& & \textbf{AWA1} &
& & \textbf{AWA2} &
\\
\textbf{Classifier}  
&  $\mathcal{Y}^S$ & $\mathcal{Y}^U$ & $H$
&  $\mathcal{Y}^S$ & $\mathcal{Y}^U$ & $H$
&  $\mathcal{Y}^S$ & $\mathcal{Y}^U$ & $H$
&  $\mathcal{Y}^S$ & $\mathcal{Y}^U$ & $H$

\\ \hline \hline
\textbf{Semantic approach} &&& &&& &&& &&& \\

SJE~\cite{akata2015evaluation} &
$59.2$ & $23.5$ & $33.6$ & 
$30.5$ & $14.7$ & $19.8$ & 
$74.6$ & $11.3$ & $19.6$ & 
$73.9$ & $8.0$ & $14.4$ \\

ALE \cite{akata2016label} &
$62.8$ & $23.7$ & $34.4$ & 
$33.1$ & $21.8$ & $26.3$ & 
$76.1$ & $16.8$ & $27.5$ & 
$81.8$ & $14.0$ & $23.9$ \\

LATEM~\cite{xian2016latent} & 
$57.3$ & $15.2$ & $24.0$ & 
$28.8$ & $14.7$ & $19.5$ & 
$71.7$ & $7.3$ & $13.3$  & 
$77.3$ & $11.5$ & $20.0$ \\

ESZSL~\cite{romera2015embarrassingly} & 
$63.8$ & $12.6$ & $21.0$ & 
$27.9$ & $11.0$ & $15.8$ & 
$75.6$ & $6.6$ & $12.1$ & 
$77.8$ & $5.9$ & $11.0$ \\

SYNC \cite{changpinyo2016synthesized} &
$70.9$ & $11.5$ & $19.8$ & 
$43.3$ & $7.9$ & $13.4$ & 
${87.3}$ & $8.9$ & $16.2$ & 
$90.5$ & $10.0$ & $18.0 $ \\

DEVISE~\cite{frome2013devise}  &
$53.0$ & $23.8$ & $32.8$ & 
$27.4$ & $16.9$ & $20.9$ & 
$68.7$ & $13.4$ & $22.4$ & 
$74.7$ & $17.1$ & $27.8 $ \\

\hline
\hline
\textbf{Generative approach} &&& &&& &&& &&& \\
SAE~\cite{kodirov2017semantic}        
& $18.0$ & $ 8.8$ & $11.8$
& $54.0$ & $ 7.8$ & $13.6$
& $77.1$ & $ 1.8$ & $ 3.5$
& $82.2$ & $ 1.1$ &	$ 2.2$
\\

f-CLSWGAN \cite{xian2017feature} & 
$57.7$ & $43.7$ & $49.7$ & 
$36.6$ & $42.6$ & $39.4$ & 
$61.4$ & $57.9$ & $59.6 $ & 
$68.9$ & $52.1 $ & $59.4$ \\

{cycle-WGAN} \cite{felix2018multi}
& $60.3$ & $46.0$ & $52.2$
& $33.1$ & $\textbf{48.3}$ & $39.2$
& $63.5$ & $56.4$ & $59.7$ 
& $  - $ & $  - $ & $  - $ \\ 

CADA-VAE~\cite{schonfeld2019cada} & 
$53.5$ & $51.6$ & $52.4$  & 
$35.7$ & $47.2$ & $40.6$  & 
$72.8$ & $57.3$ & $64.1$  & 
$75.0$ & $55.8$ & $63.9$ \\


\hline
\textbf{Combining classifiers} &&& &&& &&& &&& \\
CMT \cite{socher2013zero} & 
$49.8$ & $7.2$ & $12.6$ & 
$21.8$ & $8.1$ & $11.8$ & 
$87.6$ & $0.9$ & $1.8$ & 
${90.0}$ & $0.5$ & $1.0$ \\

DAZSL~\cite{atzmon2019adaptive}  &
$56.9$	& $47.6$ &	$51.8$ &
$37.2$	& $45.6$ &	$\textbf{41.4}$ &
$76.9$	& $54.7$ &	$63.9$ &
$-$ & $-$ & $-$ \\

\hline
\hline

{CADA-VAE} \textbf{*}  &
$57.2$ & $48.4$ & $52.4$  & 
$36.8$ & $45.1$ & $40.6$  & 
$76.6$ & $55.0$ & $64.1$  & 
$75.3$ & $55.5$ & $63.9$ \\

{MCADA-VAE} (ours)
& $55.2$ & $\textbf{52.7}$ & $\textbf{54.0}$
& $35.6$ & $47.4$ & $40.7$
& $75.2$ & $\textbf{57.3}$ & $\textbf{65.0}$
& $73.2$ & $\textbf{58.5}$ & $\textbf{65.0}$
\\
\hline

\end{tabular}
}
\end{table*}


\begin{table}[t!]
\centering
\caption{Area under the curve of seen and unseen accuracy (AUSUC).  The highlighted values per column represent the best results in each data set. The notation $\textbf{*}$ represents the results that we reproduced.}
\label{table:gzsl_aucsuac}

    \begin{tabular}{|l|c|c|c|c|}
        \hline
        \textbf{Classifier}  
        & \textbf{CUB} 
        & \textbf{SUN}
        & \textbf{AWA1 }
        & \textbf{AWA2 } \\
        \hline
        {EZSL}~\cite{romera2015embarrassingly}
        & $30.2$ & $12.8$ & $39.8$ & $ -$ \\

        {DAZSL}~\cite{atzmon2019adaptive}
        & $35.7$ & ${23.9}$ & $53.2$ & $-$ \\
        
        {f-CLSWGAN}~\cite{xian2017feature}
        & $35.5$ & $22.0$ & $46.1$ & $-$ \\
        
        {cycle-WGAN}~\cite{felix2018multi}\textbf{*}~
        & $\textbf{41.8}$ & $23.2$ & $47.3$ & $ - $\\

        {CADA-VAE}~\cite{schonfeld2019cada}\textbf{*}
        & $37.0$ & $23.6$ & $52.4$ & $52.2$ \\
        
        \hline
        \hline
        
        {MCADA-VAE} (ours)
        & $39.3$ & $\textbf{24.0}$ & $\textbf{53.2}$ & $\textbf{54.9}$\\
        \hline
     
    \end{tabular}
\end{table}

We adopted the model CADA-VAE~\cite{schonfeld2019cada} as our baseline. The code to reproduce CADA-VAE experiments from ~\cite{schonfeld2019cada} is available online\footnote{https://github.com/edgarschnfld/CADA-VAE-PyTorch}. This model is explained in Sec.~\ref{sec:data_augmentation}, where the visual encoder is a network comprising one hidden layer with $1560$ nodes, and the semantic encoder is a network consisting of one hidden layer with 1450 nodes. The visual decoder and the semantic decoder are represented by  networks with one hidden layer containing $1560$ and $660$ nodes, respectively. The latent space $\mathcal{Z}$ contains 64 dimensions. The model is optimised with Adam for 100 epochs~\cite{kingma2014adam}. As suggested in~\cite{schonfeld2019cada}. We use an adaptive rate with scheduling routine for the hyper-parameters $\gamma_{CM}$, and $\gamma_{DA}$ by $(0.044, 0.0026)$, adapted in the respective epochs $(21-75, 0-90)$.

Each of the multi-modal classifiers in~\eqref{eq:proposed_multimodal_classifier} is represented by a neural network with one linear layer transformation and an output layer of size $|\mathcal{Y}|$. As proposed in~\eqref{eq:softmax_temperature}, all these classifier networks have a softmax activation function after the linear layer. The training of these classifiers rely on multi-class cross-entropy loss and Adam optimiser~\cite{kingma2014adam}, with a learning rate of $0.001$. To alleviate the lack of unseen samples, we generated artificial samples from the semantic representation for all the benchmark data set during the training of all the classifiers. Furthermore, we calibrate the predictions with temperature scaling for GZSL models, as described in~\eqref{eq:softmax_temperature}, where the training depends on the validation set provided by Xian et. al~\cite{xian2017zero}. Then the calibration scaling is used during test time to calibrate the classification result from the softmax activation. Each classifier has a singular temperature scale.\footnote{Link to our Github repository here after review.}

\subsection{Results}
\label{sec:results}

\begin{figure*}[ht!]
    \centering
    \includegraphics[width=0.6\paperwidth]{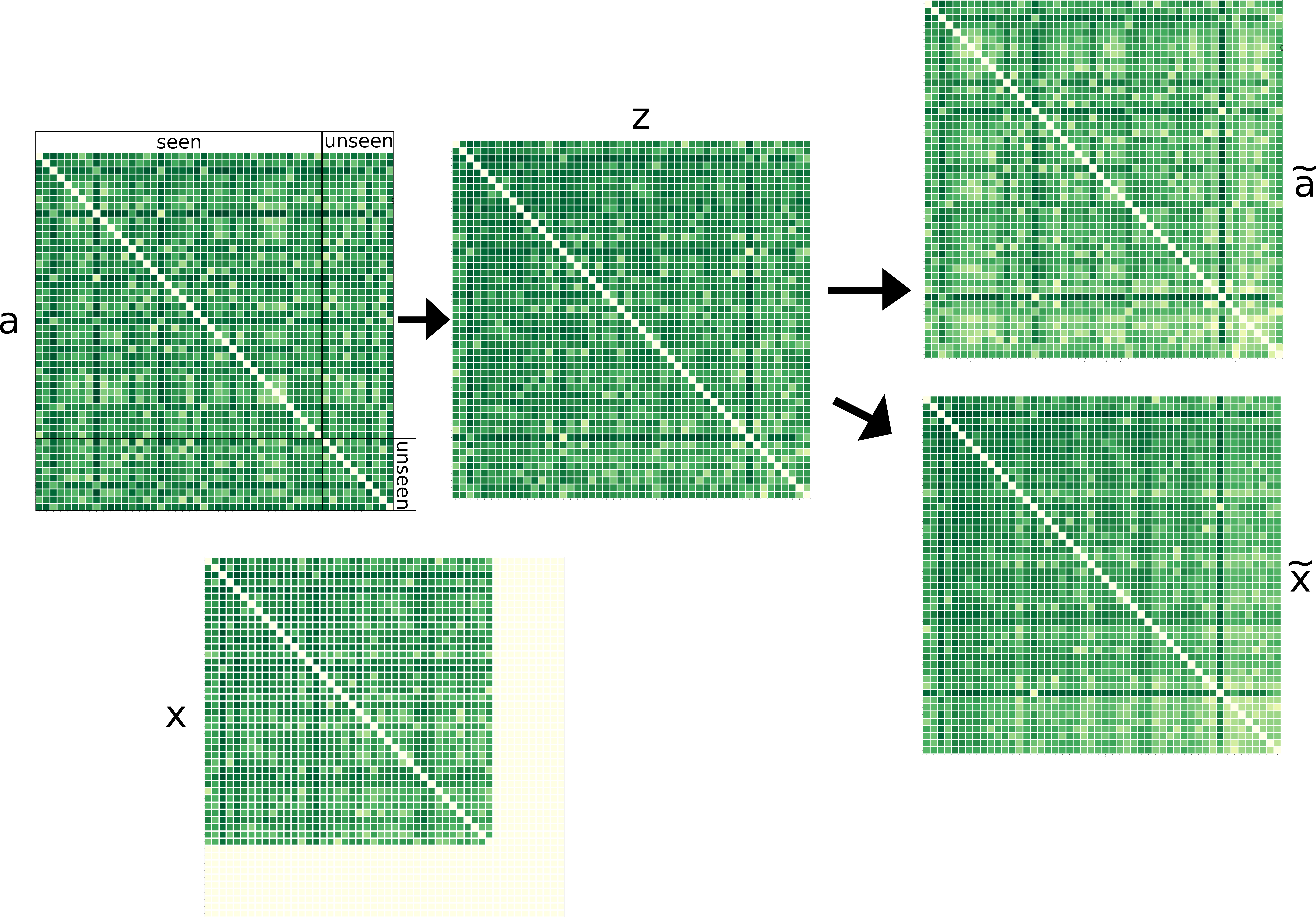}
    \caption{Pairwise distance matrices for all classes in the latent spaces of the AWA1 data set. Each matrix is divided into four quadrants: top-left (seen vs seen), top-right/bottom-left (seen vs unseen) and bottom-right (unseen vs unseen). Matrix $\mathbf{a}$ and $\tilde{\mathbf{a}}$ contains the pairwise distances for the original and reconstructed semantic representations of each class. Matrix $\mathbf{z}$ and $\tilde{\mathbf{x}}$ present the pairwise distances for the semantic representation of each class when mapped into the joint latent space and then into the reconstructed visual space. Lastly, matrix $\mathbf{x}$ contains the pairwise distances between the average visual feature of each seen class}
    \label{fig:distance_ablation}
\end{figure*}

Tables~\ref{table:gzsl_ablation} and \ref{table:gzsl_results} present numerical results in terms of unseen class accuracy $\mathcal{Y}^U$, seen class accuracy $\mathcal{Y}^S$ and harmonic mean $H$, as described in Sec.~\ref{sec:evaluation_protocol}. In Table~\ref{table:gzsl_results} we evaluate the performance of our approach, referred to as Multi-CADA-VAE (MCADA-VAE), and compare it to several models in the literature. More specifically, we show the results for the data sets CUB, SUN, AWA1 and AWA2, and compare our model to 12 GZSL methods. We define three distinct groups of GZSL approaches: the semantic approach, generative approach and models that combine disjoint domain classifiers. In the semantic approach we compare our results to SJE~\cite{akata2015evaluation}, ALE \cite{akata2016label}, LATEM~\cite{xian2016latent}, ESZSL~\cite{romera2015embarrassingly}, SYNC \cite{changpinyo2016synthesized} and DEVISE~\cite{frome2013devise}). This group focus on learning a transformation from visual to semantic representation, then the classification is based on nearest neighbour classification. For the generative approach we compare our model to SAE~\cite{kodirov2017semantic}, f-CLSWGAN \cite{xian2017feature}, {cycle-WGAN} \cite{felix2018multi} and CADA-VAE~\cite{schonfeld2019cada}). This group of models focus on learning a latent space that can be used to perform data augmentation for GZSL data sets. Finally, we compare our model to the approaches that combine the seen and unseen domain classifiers: CMT\cite{socher2013zero} and DAZSL~\cite{atzmon2019adaptive}.

Table~\ref{table:gzsl_ablation} shows an ablation study of our proposed MCADA-VAE, which is built upon the code provided in~\cite{schonfeld2019cada}. Our ablation results shows the accuracy of the classifiers trained for each modality: the joint visual/semantic embedding space $\textbf{z}$-CADA-VAE (our reproduction of CADA-VAE~\cite{schonfeld2019cada}), the reconstructed visual space $\tilde{\textbf{x}}$-CADA-VAE and the reconstructed semantic space $\tilde{\textbf{a}}$-CADA-VAE. Fig.~\ref{fig:distance_ablation} further illustrates the differences between these three spaces. Finally, we also present a naive combination of these classifiers by a simple average operation MCADA-VAE ($\tau=1$). In Sec.~\ref{sec:discussions}, we discuss the numerical results of Tables~\ref{table:gzsl_ablation} and~\ref{table:gzsl_results}, and the qualitative results of Fig.~\ref{fig:distance_ablation}.

In Table~\ref{table:gzsl_aucsuac} we show the area under the curve of seen and unseen accuracy (AUSUC) results~\cite{chao2016empirical}. We evaluate our model in terms of AUSUC for the benchmark data sets CUB, SUN, AWA1, and AWA2; and compare our results with five GZSL models: ESZSL~\cite{romera2015embarrassingly}, DAZSL~\cite{atzmon2019adaptive}, f-CLSWGAN~\cite{xian2017feature}, cycle-WGAN~\cite{felix2018multi} and CADA-VAE~\cite{schonfeld2019cada}.

\section{Discussions}\label{sec:discussions}

In Sec.~\ref{sec:ensemble_MM_classifiers} we have introduced a novel ensemble of classifiers for GZSL. Table~\ref{table:gzsl_results} shows that the proposed method MCADA-VAE outperforms other GZSL approaches in terms of unseen accuracy, $\mathcal{Y}^U$ in three out of four data sets: CUB, AWA1 and AWA2. Firstly, the comparisons in terms of unseen accuracy show that our approach improves over CADA-VAE and the previous best methods in the field in these three data sets.
Secondly, the proposed method also achieved improvements in terms of the harmonic mean for the data sets CUB, AWA1 and AWA2 concerning CADA-VAE and previous best methods in the field.

For the data set SUN, we observe that our method provides competitive performance to previous methods in the literature. Moreover, when our model is compared to our implementation of CADA-VAE on SUN, we note that we surpass it by $2.3\%$ for the unseen accuracy (from $45.1\%$ to $47.4\%$). 

The ablation study in Table~\ref{table:gzsl_aucsuac} shows that our approach is more accurate than the classifier trained in each modality (joint semantic/visual space, reconstructed visual and reconstructed semantic spaces). We also show in Table~\ref{table:gzsl_aucsuac} that the naive combination of these models, with the temperature scaling $\tau$ equal to one, does not outperform our method. These results show that temperature scaling is critical for the ensemble of classifiers.

Furthermore, Table~\ref{table:gzsl_aucsuac} shows that the proposed approach achieved solid improvement in terms of AUSUC, considering the previous state of the art. More specifically, our approach produces the highest AUSUC in three out of the four data sets (SUN, AWA1, and AWA2), and also improves over CADA-VAE on all four data sets.

We largely attribute the improved performance of our method to the additional information that is available across our ensemble of classifiers. Learning transformations between modalities can result in both losses and gains in the information. We assume that information is lost when representation capacity causes distinct data points to be mapped to the same location in a learned space. Alternatively, it is also possible to enrich a joint embedding space with information from multiple sources, and for this information to then be represented within the mappings from each of those modalities. Fig.~\ref{fig:distance_ablation} illustrates this concept with pairwise distance matrices generated for each embedding space in our proposed method. Here, each of the learning spaces is derived from the same visual and semantic representations. However, we observe that the spatial distribution of classes is different for each learned space. By giving our classifiers access to each of these spaces, our approach can take advantage of information that is not present across each of these different representations.

\section{Conclusions and Future Work}\label{sec:conclusions}

In this paper, we introduce an approach that combines multi-modal GZSL classifiers. In particular, we extend our approach from the CADA-VAE and show that the multiple spaces optimised by this model can contribute to a powerful ensemble of classifiers. Furthermore, we showed that the temperature scaling that has been widely used for calibrating confidences of classifiers can also be used to mitigate model selection in ensemble classification. Our experimental results provide evidence for these contributions and demonstrate that our approach achieves substantial improvements in almost all of the benchmarks adopted in this field.
Specifically, our proposed method achieved state-of-the-art H-mean results for CUB, AWA1 and AWA2. In terms of unseen accuracy, our method also outperforms all previous methods for the CUB, AWA1, and AWA2 datasets. In particular, our results are substantially better than the state-of-the-art for CUB and AWA2. Moreover, our model achieves state-of-the-art results in terms of AUSUC for SUN, AWA1 and AWA2.

In Sec.~\ref{sec:discussions}, we discussed how our ensemble of multiple classifiers can combine complementary information from multiple embedding spaces to improve upon single classifier baselines. We believe that our initial result in this area warrants further study in the context of GZSL and for enhancing other classification domains. Furthermore, we have shown that temperature scaling can be successfully applied to an ensemble of GZSL models. As such, further research is needed to study whether temperature scaling can be applied more generally to the ensemble of classifiers. Lastly, we believe that the large size of the SUN data set makes it particularly challenging in the context of GZSL. As such, there is still room for new approaches that can deal with this type of challenge.

\section{Acknowledgement}\label{sec:acknnowledgement}

Supported by Australian Research Council through grants \textit{DP180103232}, \textit{CE140100016} and \textit{FL130100102}. We would like to acknowledge the donation of a TitanXp by Nvidia.

{\small
\bibliographystyle{ieee}
\bibliography{egbib}
}

\end{document}